\documentclass{article}

\usepackage{graphicx} 

\usepackage{natbib}

\usepackage{algorithm}
\usepackage{algorithmic}

\usepackage{amsmath}
\usepackage{amssymb}
\usepackage{appendix}
\usepackage{amsthm}
\usepackage{ifthen}

\usepackage{hyperref}

\usepackage[accepted]{icml2011}

\newboolean{longversion}
\setboolean{longversion}{true}

\makeatletter
\newtheorem*{rep@theorem}{\rep@title}
\newcommand{\newreptheorem}[2]{%
\newenvironment{rep#1}[1]{%
 \def\rep@title{#2 \ref{##1}}%
 \begin{rep@theorem}}%
 {\end{rep@theorem}}}
\makeatother

\newtheorem{theorem}{Theorem}
\newreptheorem{theorem}{Theorem}
\newtheorem{definition}{Definition}
\newreptheorem{definition}{Definition}

\newcommand{\innerprod}[3][\defmu]{{\left\langle {#2},{#3} \right\rangle}_{#1}}
\newcommand{\norm}[2][\defmu]{{\left\lVert {#2} \right\rVert}_{#1}}
\newcommand{\functional}[3][\defmu]{{\mathcal{#2}}_{#1} [#3]}
\newcommand{\risk}[2][\defmu]{\functional[#1]{R}{#2}}

\DeclareMathOperator*{\argmax}{arg\,max}
\DeclareMathOperator*{\argmin}{arg\,min}

\icmltitlerunning{Generalized Boosting Algorithms for
Convex Optimization}

\begin{document} 

\twocolumn[
\icmltitle{Generalized Boosting Algorithms for Convex Optimization}

\icmlauthor{Alexander Grubb}{agrubb@cmu.edu}
\icmlauthor{J. Andrew Bagnell}{dbagnell@ri.cmu.edu}
\icmladdress{School of Computer Science,
             Carnegie Mellon University,
             Pittsburgh, PA 15213 USA
            }

\icmlkeywords{}

\vskip 0.3in
]

\begin{abstract}

  Boosting is a popular way to derive powerful
  learners from simpler hypothesis classes.
  Following previous work \cite{mason:99,friedman:00} on
  general boosting frameworks, we analyze gradient-based
  descent algorithms for boosting with respect to any
  convex objective and introduce a new measure of
  weak learner performance into this setting which generalizes
  existing work.
  We present the weak to strong learning guarantees
  for the existing gradient boosting work for strongly-smooth,
  strongly-convex
  objectives under this new measure of performance,
  and also demonstrate that this work fails for
  non-smooth objectives.
  To address this issue, we present new
  algorithms which extend this boosting approach to arbitrary
  convex loss functions and give corresponding weak to
  strong convergence results.
  In addition, we demonstrate experimental
  results that support our analysis and demonstrate the need
  for the new algorithms we present.

\end{abstract}


\section{Introduction}
\label{sec:intro}

Boosting \cite{schapire:02} is a versatile meta-algorithm for
combining together multiple simple hypotheses, or weak learners,
to form a single complex hypothesis with superior performance.  The power
of this meta-algorithm lies in its ability to craft hypotheses which
can achieve arbitrary performance on training data using only
weak learners that perform marginally better than random.  This
\emph{weak to strong} learning guarantee is a critical feature of boosting.

To date, much of the work on boosting has focused on optimizing
the performance of this meta-algorithm with respect to specific
loss functions and problem settings.  The AdaBoost algorithm \cite{freund:97}
is perhaps the most well known and most successful of these.
AdaBoost focuses specifically on the task of classification via the
minimization of the exponential loss by boosting weak binary classifiers
together, and can be shown to be near optimal in this setting.
Looking to extend upon the success of AdaBoost, related algorithms have
been developed for other domains, such as RankBoost \cite{freund:03} and
mutliclass extensions to AdaBoost \cite{mukherjee:10}.
Each of these algorithms provides both strong theoretical
and experimental results for their specific domain, including
corresponding weak to strong learning guarantees,
but extending boosting to these and other new settings is
non-trivial.

Recent attempts have been successful at generalizing the boosting
approach to certain broader classes of problems, but their focus is also
relatively restricted.  Mukherjee and Schapire \yrcite{mukherjee:10}
present a general theory of boosting for multiclass classification
problems, but their analysis is restricted to the multiclass setting.
Zheng et al. \yrcite{zheng:10} give a boosting method which
utilizes the second-order Taylor approximation of the objective
to optimize smooth, convex losses.
Unfortunately, the corresponding convergence result
for their algorithm does not
exhibit the typical weak to strong guarantee seen in
boosting analyses and their results apply only to weak learners
which solve the weighted squared regression problem.

Other previous work on providing general algorithms for boosting
has shown that an intuitive link between algorithms like AdaBoost and
gradient descent exists \cite{mason:99,friedman:00}, and that
many existing boosting algorithms can be reformulated
to fit within this gradient boosting framework.  Under this view,
boosting algorithms are seen as performing a modified gradient descent
through the space of all hypotheses, where the gradient is calculated
and then used to find the weak learner which will provide the
best descent direction.

In the case of smooth convex functionals, Mason et al. \yrcite{mason:99}
give a proof of eventual convergence for this previous work, but no rates
of convergence are given.
Additionally, convergence rates of these algorithms have been
analyzed for the case of smooth convex functionals \cite{ratsch:02}
and for specific potential functions used in classification \cite{duffy:00}
under the traditional PAC weak learning setting.

Our work aims to rigorously define the mathematics underlying
this connection and show how standard boosting notions such
as that of weak learner performance can be extended to the
general case.  Using this foundation, we will present
weak to strong learning results for the existing gradient
boosting algorithm \cite{mason:99,friedman:00} for the special
case of smooth convex objectives under our more general setting.

Furthermore, we will also demonstrate that this existing algorithm can fail
to converge on non-smooth objectives, even in finite dimensions.  To rectify this issue,
we present new algorithms which do have corresponding strong convergence
guarantees for all convex objectives, and demonstrate experimentally
that these new algorithms often outperform the existing algorithm in practice.

Our analysis is modeled after existing work on gradient
descent algorithms for optimizing over vector spaces.
For convex problems
standard gradient descent algorithms are known to provide
good convergence results \cite{zinkevich:03,boyd:04,hazan:06}
and are widely applicable.  However, as detailed above,
the modified gradient descent procedure which
corresponds to boosting does not directly follow the gradient,
instead selecting a descent direction from a restricted set of
allowable search directions.  This \emph{restricted gradient descent}
procedure requires new extensions to the previous work on
gradient descent optimization algorithms.

A related form of gradient descent with gradient errors
has previously been studied in the analysis of budgeted learning
\cite{sutskever:09}, and general results related to gradient projection
errors are given in the literature.  While these results apply
to the boosting setting, they lack any kind of weak to strong
guarantee.  Conversely, we are primarily interested in
studying what algorithms and assumptions are needed to overcome projection
error and achieve strong final performance even in the face of
mediocre weak learner performance.

The rest of the paper is as follows.  We first explicitly detail
the Hilbert space of functions and various operations within this Hilbert space.
Then, we discuss how to quantify the performance of a weak learner in terms
of this vector space.  Following that, we present theoretical weak to strong
learning guarantees for both the existing and our new algorithms.  Finally
we provide experimental results comparing all algorithms discussed on a
variety of tasks.

\section{$L^2$ Function Space}
\label{sec:space}
Previous work \cite{mason:99,friedman:00} has presented the theory underlying
function space gradient descent in a variety of ways, but never in a form
which is convenient for convergence analysis.  Recently,
Ratliff \yrcite{ratliff:09b} proposed the $L^2$ function space as a natural
match for this setting.  This representation as a vector space is particularly
convenient as it dovetails nicely with the analysis of gradient descent
based algorithms.  We will present here the Hilbert space of functions
most relevant to functional gradient boosting, but the later convergence analysis for
restricted gradient descent algorithms can be generalized to any Hilbert space.

Given a measurable input set $\mathcal{X}$, an output
vector space $\mathcal{V}$, and measure $\mu$,
the function space $L^2(\mathcal{X},\mathcal{V},\mu)$ is the set of all equivalence
classes of functions
$f : \mathcal{X} \rightarrow \mathcal{V}$ such that the Lebesgue integral
\begin{equation}
\label{eq:lebesgue}
\int_{\mathcal{X}} \norm[\mathcal{V}]{f(x)}^2\, d\mu
\end{equation}
is finite.  We will specifically consider the special case
where $\mu$ is a probability measure $P$ with density function $p(x)$,
so that (\ref{eq:lebesgue}) is equivalent to $\mathbb{E}_{P}[\norm{f(x)}^2]$.

This Hilbert space has a natural inner product and norm:
\begin{align*}
  \innerprod[P]{f}{g} &= \int_{\mathcal{X}} \innerprod[\mathcal{V}]{f(x)}{g(x)}\, p(x)\, dx\\
  &= \mathbb{E}_{P}[\innerprod[\mathcal{V}]{f(x)}{g(x)}]\\
  \norm[P]{f}^2 &= \innerprod[P]{f}{f}\\ 
  &= \mathbb{E}_{P}[\norm[\mathcal{V}]{f(x)}^2].
\end{align*}

We parameterize these operations by $P$ to denote their
reliance on the underlying data distribution.
In the case of the empirical probability distribution $\hat{P}$
these quantities are simply the corresponding
empirical expected value.  For example, the inner product becomes
\[
\innerprod[\hat{P}]{f}{g} = \frac{1}{N} \sum_{n=1}^N \innerprod[\mathcal{V}]{f(x_n)}{g(x_n)}
\]

In order to perform gradient descent over such a space,
we need to compute the gradient of functionals over said space.
We will use the standard definition of a \emph{subgradient} to allow for
optimization of non-smooth functions.  Define $\nabla \risk{f}$
to be a subgradient iff:
\[
\risk{f} \ge \risk{g} + \innerprod[P]{f - g}{\nabla \risk{f}}
\]
Here $\nabla \risk{f}$ is a (function space) subgradient of
the functional $\mathcal{R} : L^2(P) \rightarrow \mathbb{R}$
at $f$.  Using this definition, these subgradients are
straightforward to compute for a number of functionals.

For example, for the point-wise loss
over a set of training examples,
\[
\functional[\textrm{emp}]{R}{f} = \frac{1}{N} \sum_{n=1}^N l(f(x_n), y_n)
\]
the subgradients in $L^2(\mathcal{X},\mathcal{V},\hat{P})$ are the set:
\[
\nabla \functional[\textrm{emp}]{R}{f} = \left\{ g \ | \ g(x_n) \in (\nabla_1 l)(f(x_n), y_n) \right\}
\]
where $(\nabla_1 l)(f(x_n),y_n)$ is the set of subgradients of the pointwise loss $l$
with respect to $f(x_n)$.  For differentiable $l$, this is
just the partial derivative of $l$ with respect to input $f(x_n)$.

Similarly the expected loss,
\[
\functional{R}{f} = \mathbb{E}_{P}[\mathbb{E}_{\mathcal{Y}}[l(f(x),y)]],
\]
has the following subgradients in  $L^2(\mathcal{X},\mathcal{V},P)$:
\[
\nabla \functional{R}{f} = \left\{ g \ | \ g(x) \in \mathbb{E}_{\mathcal{Y}} [(\nabla_1 l)(f(x), y)] \right\}.
\]


\section{Restricted Gradient Descent}
\label{sec:projgrad}

We now outline the gradient-based view of boosting \cite{mason:99,friedman:00}
and how it relates to gradient descent.
In contrast to the standard gradient descent algorithm,
boosting is equivalent to what we will call
the \emph{restricted gradient descent} setting,
where the gradient is not followed directly,
but is instead replaced by another search
direction from a set of allowable descent directions.
We will refer to this set of allowable directions as
the \emph{restriction set}.

From a practical standpoint, a projection step is necessary 
when optimizing over function space
because the functions representing the gradient directly
are computationally difficult to manipulate and do not generalize to new
inputs well.
In terms of the connection to boosting,
the restriction set corresponds directly to the
set of hypotheses generated by a weak learner.

We are primarily interested in two aspects of this restricted
gradient setting:
first, appropriate ways to find the best allowable
direction of descent, and second, a means of quantifying
the performance of a restriction set.  Conveniently, the function
space view of boosting provides a simple geometric
explanation for these concerns.

Given a gradient $\nabla$ and candidate direction $h$,
the closest point $h'$ along $h$ can be found using vector projection:
\begin{equation}
  \label{eq:projection}
  h' = \frac{\innerprod{\nabla}{h}}{\norm{h}^2} h
\end{equation}

Now, given a set of possible descent directions $\mathcal{H}$
the vector $h^*$ which minimizes the resulting projection error
(\ref{eq:projection}) also maximizes the projected length:
\begin{equation}
  \label{eq:projinnerprod}
  h^* = \underset{h \in \mathcal{H}}{\argmax}\ \frac{\innerprod{\nabla}{h}}{\norm{h}}.
\end{equation}
This is a generalization of the projection operation in Mason et al. \yrcite{mason:99}
to functions other than classifiers.

For the special case when $\mathcal{H}$ is closed under scalar multiplication, one can
instead find $h^*$ by directly minimizing the distance between $\nabla$ and $h^*$,
\begin{equation}
  \label{eq:projnorm}
  h^* = \underset{h \in \mathcal{H}}{\argmin}\ \norm{\nabla - h}^2
\end{equation}
thereby reducing the final projected distance found using (\ref{eq:projection}).  This projection
operation is equivalent to the one given by Friedman \yrcite{friedman:00}.

These two projection methods provide relatively simple ways to
search over any restriction set for the `best' descent
direction.  The straightforward algorithm \cite{mason:99,friedman:00} for peforming
restricted gradient descent which uses these projection
operations is given in Algorithm~\ref{alg:naive}.

\begin{algorithm}[tb]
  \caption{Naive Gradient Projection Algorithm}
  \label{alg:naive}
  \begin{algorithmic}
    \STATE {\textbf{Given:}} starting point $f_0$, step size schedule $\{\eta_t\}_{t=1}^T$
    \FOR{$t = 1,\ldots,T$}
    \STATE Compute subgradient $\nabla_t \in \nabla \functional{R}{f}$.
    \STATE Project $\nabla_t$ onto hypothesis space $\mathcal{H}$, finding nearest direction $h^*$.
    \STATE Update $f$: $f_t \leftarrow f_{t-1} - \eta_t \frac{\innerprod{h^*}{\nabla_t}}{\norm{h^*}^2} h^*$.
    \ENDFOR
  \end{algorithmic}
\end{algorithm}

In order to analyze the restricted gradient descent
algorithms, we need a way quantify the relative strength of
a given restriction set.  A guarantee on the
performance of each projection step, typically referred to
in the traditional boosting literature as the \emph{edge}
of a given weak learner
is crucial to the convergence analysis of restricted
gradient algorithms.

For the projection which maximizes the inner product as in
(\ref{eq:projinnerprod}),
we can use the generalized geometric notion of angle to
bound performance by requiring that
\[
\innerprod{\nabla}{h} \ge \cos \theta \norm{\nabla}\norm{h}
\]
while the equivalent requirement for the norm-based projection
in (\ref{eq:projnorm}) is
\[
\norm{\nabla - h}^2 \le (1 - (\cos \theta)^2) \norm{\nabla}^2.
\]

Parameterizing by $\cos \theta$, we can now concisely define the
performance potential of a restricted set of search directions,
which will prove useful in later analysis.

\begin{definition}
  \label{def:edge}
  A restriction set $\mathcal{H}$ has \emph{edge} $\gamma$ if
  for every projected gradient $\nabla$ there exists a vector $h \in \mathcal{H}$
  such that either $\innerprod{\nabla}{h} \ge \gamma \norm{\nabla}\norm{h}$
  or $\norm{\nabla - h}^2 \le (1 - \gamma^2) \norm{\nabla}^2$.
\end{definition}

This definition of edge is parameterized by $\gamma \in [0,1]$,
with larger values of edge corresponding to lower projection
error and faster algorithm convergence.


\subsection{Relationship to Previous Boosting Work}

Though these projection operations apply to
any $L^2$ hypothesis set,
they also have convenient interpretations when it comes to
specific function classes traditionally used
as weak learners in boosting.

For a classification-based weak learner with outputs
in $\{-1,+1\}$
and an optimization over single output functions
$f : \mathcal{X} \rightarrow \mathbb{R}$,
projecting as in (\ref{eq:projinnerprod}) is equivalent
to solving the weighted classification problem
over examples $\{x_n, \operatorname{sgn}(\nabla(x_n))\}_{n=1}^N$
and weights $w_n = |\nabla(x_n)|$.

The projection via norm minimization in (\ref{eq:projnorm})
is equivalent to solving the regression problem
\[
h^* = \underset{h \in \mathcal{H}}{\argmin}\ \frac{1}{N} \sum_{n=1}^N \norm{\nabla(x_n) - f(x_n)}^2
\]
using the gradient outputs as regression targets.

Similarly,
our notion of weak learner performance in Definition \ref{def:edge}
can be related to previous work.  Like our measure of edge
which quantifies performance over the trivial hypothesis
$h(x) = 0, \forall x$, previous work has used similar
quantities which capture the advantage over baseline
hypotheses.

For weak learners
which are binary classifiers, as is the case in AdaBoost \cite{freund:97},
there is an equivalent notion of edge
which refers to the improvement in performance over predicting
randomly.  We can show that Definition \ref{def:edge}
is an equivalent measure:

\begin{theorem}
  \label{theorem:ada-equivalence}
  For a weak classifier space $\mathcal{H}$
  with outputs in $\{-1,+1\}$, the following statements are
  equivalent: (1) $\mathcal{H}$ has edge $\gamma$ for some
  $\gamma > 0$, and
  (2) for any non-negative weights $w_n$ over training data
  $x_n$, there is a classifier $h \in \mathcal{H}$ which
  achieves an error of at most
  $(\frac{1}{2} - \frac{\delta}{2}) \sum_n w_n$
  for some $\delta > 0$.
\end{theorem}

A similar result can be shown for more recent work on multiclass
weak learners \cite{mukherjee:10} when optimizing over functions
with multiple outputs $f : \mathcal{X} \rightarrow \mathbb{R}^k$:

\begin{theorem}
  \label{theorem:multiclass-equivalence}
  For a weak multiclass classifier space $\mathcal{H}$
  with outputs in $\{1,\ldots,K\}$, let the modified hypothesis space
  $\mathcal{H}'$ contain a hypothesis $h': \mathcal{X} \rightarrow \mathbb{R}^K$
  for each $h \in \mathcal{H}$
  such that $h'(x)_k = 1$ if $h(x) = k$ and $h'(x) = -\frac{1}{K-1}$ otherwise.
  Then, the following statements are
  equivalent: (1) $\mathcal{H'}$ has edge $\gamma$ for some $\gamma > 0$, and
  (2) $\mathcal{H}$ satisfies the performance over baseline requirements
  detailed in Theorem 1 of \cite{mukherjee:10}.
\end{theorem}

Proofs and more details on these equivalences can be found in
\ifthenelse{\boolean{longversion}}
{Appendix \ref{app:equiv}.}
{the extended version of the paper \cite{withproofs}.}

\section{Convergence Analysis}

We now focus on analyzing the behavior of variants of the basic restricted gradient
descent algorithm shown in Algorithm \ref{alg:naive} on problems of the
form:
\[
\min_{f \in \mathcal{F}} \functional[\textrm{emp}]{R}{f},
\]
where allowable descent directions are taken from some
restriction set $\mathcal{H} \subset \mathcal{F}$.

In line with previous boosting work, we will specifically consider
cases where the edge requirement in Definition \ref{def:edge}
is met for some $\gamma$, and seek convergence results where
the empirical objective $\functional[\textrm{emp}]{R}{f_t}$ approaches the optimal
training performance $\min_{f \in \mathcal{F}} \functional[\textrm{emp}]{R}{f}$.
This work does not attempt to analyze the convergence of the true risk,
$\functional{R}{f}$.

While we consider $L^2$ function space specifically,
the convergence analysis presented can be extended to optimization
over any Hilbert space using restricted gradient descent.

\subsection{Smooth Convex Optimization}

An earlier result showing $O((1 - \frac{1}{C})^T)$ convergence
of the objective to optimality for smooth functionals is given by R\"atsch,
et al. \cite{ratsch:02} using results from the optimization literature
on coordinate descent.  Alternatively, this gives a $O(\log(\frac{1}{\epsilon}))$
result for the number of iterations required to achieve error $\epsilon$.
Similar to our result, this work relies on the smoothness
of the objective as well as the weak learner performance,
but uses the more restrictive notion of edge from previous boosting
literature specifically tailored to PAC weak learners (classifiers).
This previous result also has an additional dependence on the number of
weak learners and number of training examples.

We will now give a generalization of the result in \cite{ratsch:02}
which uses our more general definition of weak learner edge.
The convergence analysis of Algorithm \ref{alg:naive} relies
on two critical properties of the objective functional $\mathcal{R}$.

A functional $\mathcal{R}$ is \emph{$\lambda$-strongly convex} if
$\forall f,f' \in \mathcal{F}$:
\[
\functional{R}{f'} \ge \functional{R}{f}
+ \innerprod{\nabla \functional{R}{f}}{f' - f}
+ \frac{\lambda}{2} \norm{f' - f}^2
\]
for some $\lambda > 0$,
and \emph{$\Lambda$-strongly smooth} if
\[
\functional{R}{f'} \le \functional{R}{f}
+ \innerprod{\nabla \functional{R}{f}}{f' - f}
+ \frac{\Lambda}{2} \norm{f' - f}^2
\]
for some $\Lambda > 0$.
Using these two properties, we can now derive a
convergence result for unconstrained optimization
over smooth functions.

\begin{theorem}[Generalization of Theorem 4 in \cite{ratsch:02}]
  \label{theorem:smooth-empirical}
  Let $\mathcal{R}_{\textrm{emp}}$ be a $\lambda$-strongly convex and
  $\Lambda$-strongly smooth functional over
  $L^2(\mathcal{X},\hat{P})$ space.
  Let $\mathcal{H} \subset L^2$ be a restriction set
  with edge $\gamma$.
  Let $f^* = {\argmin}_{f \in \mathcal{F}} \functional[\textrm{emp}]{R}{f}$.
  Given a starting point $f_0$ and
  step size $\eta_t = \frac{1}{\Lambda}$,
  after $T$ iterations of Algorithm \ref{alg:naive} we have:
  \[
  \functional[\textrm{emp}]{R}{f_T} - \functional[\textrm{emp}]{R}{f^*}
  \le (1 - \frac{\gamma^2 \lambda}{\Lambda})^T
  (\functional[\textrm{emp}]{R}{f_0} - \functional[\textrm{emp}]{R}{f^*}).
  \]
\end{theorem}

The result above holds for the fixed step size $\frac{1}{\Lambda}$
as well as for step sizes found using a line search along the descent
direction.  The analysis uses the strong smoothness requirement to
obtain a quadratic upper bound on the function and then makes guaranteed
progress by selecting the step size which minimizes this bound,
with larger gains made for larger values of $\gamma$.
A complete proof is provided in
\ifthenelse{\boolean{longversion}}
{Appendix \ref{app:smooth}.}
{the extended version of the paper \cite{withproofs}.}

Theorem \ref{theorem:smooth-empirical} gives, for strongly smooth objective
functionals, a convergence rate of $O((1 - \frac{\gamma^2 \lambda}{\Lambda})^T)$.
This is very similar to the $O((1 - 4 \gamma^2)^{\frac{T}{2}})$ convergence of
AdaBoost \cite{freund:97}, with both requiring $O(\log(\frac{1}{\epsilon}))$
iterations to get performance within $\epsilon$ of optimal.
While the AdaBoost result generally provides tighter bounds,
this relatively naive method of gradient projection is able to obtain
reasonably competitive convergence results while being applicable
to a much wider range of problems.  This is expected, as the proposed method
derives no benefit from loss-specific optimizations and can
use a much broader class of weak learners.  This comparison is a common
scenario within optimization: while highly specialized
algorithms can often perform better on specific problems, general
solutions often obtain equally impressive results, albeit less
efficiently, while requiring much less effort to implement.

Unfortunately, the naive approach to restricted gradient descent
breaks down quickly in more general cases such as non-smooth objectives.
Consider the following
example objective over two points $x_1, x_2$:
$\mathcal{R}[f] = 2 |f(x_1)| + |f(x_2)|$.
Now consider the hypothesis set $h \in \mathcal{H}$ such that
either $h(x_1) \in \{-1,+1\}$ and $h(x_2) = 0$ or
$h(x_1) = 0$ and $h(x_2) \in \{-1,+1\}$.
The algorithm will always select $h^*$ such that $h^*(x_2) = 0$
when projecting gradients from the example objective, giving
a final function with perfect performance on $x_1$ and arbitrarily
poor unchanged performance on $x_2$.  Even if the loss on training
point $x_2$ is substantial, the naive algorithm will not
correct it.

An algorithm which
only ever attempts to project subgradients of $\mathcal{R}$,
such as Algorithm \ref{alg:naive}, will not be able to obtain strong
performance results for cases like these.  The algorithms in the next
section overcome this obstacle by projecting modified versions of
the subgradients of the objective at each iteration.

\subsection{General Convex Optimization}

For the convergence analysis of general convex functions
we now switch to analyzing the average optimality gap:
\[
\frac{1}{T} \sum_{t=1}^T [\functional{R}{f_t} - \functional{R}{f^*}] ,
\]
where $f^* = \underset{f \in \mathcal{F}}{\argmin}\ \sum_{t=1}^T \functional{R}{f}$
is the fixed hypothesis which minimizes loss.

By showing that the average optimality gap
approaches 0 as $T$ grows large, for decreasing step sizes,
it can be shown that the optimality gap
$\functional{R}{f_t} - \functional{R}{f^*}$ also approaches 0.

This analysis is similar to the standard no-regret online learning approach,
but we restrict our analysis to the case when $\mathcal{R}_t = \mathcal{R}$.
This is because the true online setting typically involves receiving a new
dataset at every time $t$, and hence a different data distribution
$\hat{P}_t$, effectively
changing the underlying $L^2$ function space at every time step,
making comparison of quantities at different time steps difficult in the
analysis.
The convergence analysis
for the online case is beyond the scope of this paper
and is not presented here.

The convergence results to follow are similar to previous convergence
results for the standard gradient descent setting \cite{zinkevich:03,hazan:06},
but with a number of additional error terms due to the gradient
projection step.  Sutskever \yrcite{sutskever:09} has previously studied
the convergence of gradient descent with gradient projection errors
using an algorithm similar to Algorithm \ref{alg:naive},
but the analysis does not focus on the weak to strong learning guarantee we seek.
In order to obtain this guarantee we now present two new algorithms.


\begin{algorithm}[tb]
  \caption{Repeated Gradient Projection Algorithm}
  \label{alg:repeat}
  \begin{algorithmic}
    \STATE {\textbf{Given:}} starting point $f_0$, step size schedule $\{\eta_t\}_{t=1}^T$
    \FOR{$t = 1,\ldots,T$}
    \STATE Compute subgradient $\nabla_t \in \nabla \functional{R}{f}$.
    \STATE Let $\nabla' = \nabla_t$, $h^* = 0$.
    \FOR{$k = 1,\ldots,t$}
    \STATE Project $\nabla'$ onto hypothesis space $\mathcal{H}$, finding nearest direction $h^*_k$.
    \STATE $h^* \leftarrow h^* + \frac{\innerprod{h^*_k}{\nabla'}}{\norm{h^*_k}^2} h^*_k$.
    \STATE $\nabla' \leftarrow \nabla' - h^*_k$.
    \ENDFOR
    \STATE Update $f$: $f_t \leftarrow f_{t-1} - \eta_t h^*$.
    \ENDFOR
  \end{algorithmic}
\end{algorithm}

Our first general convex solution, shown in Algorithm \ref{alg:repeat},
overcomes this issue by using a meta-boosting strategy.  At each iteration $t$
instead of projecting the gradient $\nabla_t$ onto a single hypothesis $h^*$, we use the
naive algorithm to construct $h^*$ out of a small number of restricted steps, optimizing over
the distance $\norm{\nabla_t - h^*}^2$.  By increasing the number of weak learners
trained at each iteration over time, we effectively decrease the gradient projection
error at each iteration.  As the average projection error approaches 0, the performance
of the combined hypothesis approaches optimal.
We now give convergence results for this algorithm for both strongly
convex and convex functionals.

\begin{theorem}
  \label{theorem:repeat-empirical-strong-convex}
  Let $\mathcal{R}_{\textrm{emp}}$ be a $\lambda$-strongly convex
  functional over $\mathcal{F}$.
  Let $\mathcal{H} \subset \mathcal{F}$ be a restriction set
  with edge $\gamma$.  Let $\norm[\hat{P}]{\nabla \functional{R}{f}} \le G$.
  Let $f^* = {\argmin}_{f \in \mathcal{F}} \functional[\textrm{emp}]{R}{f}$.
  Given a starting point $f_0$ and
  step size $\eta_t = \frac{2}{\lambda t}$, after $T$ iterations of
  Algorithm~\ref{alg:repeat} we have:
  \[
  \frac{1}{T} \sum_{t=1}^T [\functional[\textrm{emp}]{R}{f_t} - \functional[\textrm{emp}]{R}{f^*}] \le
  \frac{G^2}{\lambda T} (1 + \ln T + \frac{1 - \gamma^2}{\gamma^2}).
  \]
\end{theorem}

The proof
\ifthenelse{\boolean{longversion}}
{(Appendix \ref{app:general})}
{\cite{withproofs}}
relies on the fact that as the number of iterations increases,
our gradient projection error approaches 0 at the rate given in Theorem \ref{theorem:smooth-empirical},
causing the behavior of Algorithm \ref{alg:repeat} to approach the standard gradient descent algorithm.
The additional error term in the result is a bound on the geometric series 
describing the errors introduced at each time step.

\begin{theorem}
\label{theorem:repeat-empirical}
Let $\mathcal{R}_{\textrm{emp}}$ be a convex
functional over $\mathcal{F}$.
Let $\mathcal{H} \subset \mathcal{F}$ be a restriction set
with edge $\gamma$.  Let $\norm[\hat{P}]{\nabla \functional{R}{f}} \le G$
and $\norm[\hat{P}]{f} \le F$ for all $f \in \mathcal{F}$.
Let $f^* = {\argmin}_{f \in \mathcal{F}} \functional[\textrm{emp}]{R}{f}$.
Given a starting point $f_0$ and
step size $\eta_t = \frac{1}{\sqrt{t}}$, after $T$ iterations of
Algorithm \ref{alg:repeat} we have:
\[
\frac{1}{T} \sum_{t=1}^T [\functional[\textrm{emp}]{R}{f_t} - \functional[\textrm{emp}]{R}{f^*}] \le
  \frac{F^2}{2 \sqrt{T}} + \frac{G^2}{\sqrt{T}} + 2 F G \frac{1-\gamma^2}{\gamma^2}.
\]
\end{theorem}

Again, the result is similar to the standard gradient descent result, with an
added error term dependent on the edge $\gamma$.

An alternative version of the repeated projection algorithm allows for a variable
number of weak learners to be trained at each iteration.  An accuracy threshold for
each gradient projection can be derived given a desired accuracy for the final
hypothesis, and this threshold can be used to train weak learners at each iteration
until the desired accuracy is reached.

\begin{algorithm}[tb]
\caption{Residual Gradient Projection Algorithm}
\label{alg:residual}
\begin{algorithmic}
\STATE {\textbf{Given:}} starting point $f_0$, step size schedule $\{\eta_t\}_{t=1}^T$
\STATE Let $\Delta = 0$.
\FOR{$t = 1,\ldots,T$}
\STATE Compute subgradient $\nabla_t \in \nabla \functional{R}{f}$.  $\Delta \leftarrow \Delta + \nabla_t$.
\STATE Project $\Delta$ onto hypothesis space $\mathcal{H}$, finding nearest direction $h^*$.
\STATE Update $f$: $f_t \leftarrow f_{t-1} - \eta_t \frac{\innerprod{h^*}{\Delta}}{\norm{h^*}^2} h^*$.
\STATE Update residual: $\Delta \leftarrow \Delta - \frac{\innerprod{h^*}{\Delta}}{\norm{h^*}^2} h^*$
\ENDFOR
\end{algorithmic}
\end{algorithm}

Algorithm \ref{alg:residual} gives a second method for optimizing over convex objectives.
Like the previous approach, the projection error at each time step is used again in projection,
but a new step is not taken immediately to decrease the projection error.
Instead, this approach keeps track of the residual error left over after
projection and includes this error in the next projection step.  This forces the projection
steps to eventually account for past errors, preventing the possibility of systematic
error being adversarially introduced through the weak learner set.

As with Algorithm \ref{alg:repeat}, we can derive similar convergence
results for strongly-convex and general convex functionals for this new residual-based
algorithm.

\begin{theorem}
\label{theorem:residual-empirical-strong-convex}
Let $\mathcal{R}_{\textrm{emp}}$ be a $\lambda$-strongly convex
functional over $\mathcal{F}$.
Let $\mathcal{H} \subset \mathcal{F}$ be a restriction set
with edge $\gamma$.  Let $\norm[\hat{P}]{\nabla \functional{R}{f}} \le G$.
Let $f^* = {\argmin}_{f \in \mathcal{F}} \functional[\textrm{emp}]{R}{f}$.
Let $c = \frac{2}{\gamma^2}$.
Given a starting point $f_0$ and
step size $\eta_t = \frac{1}{\lambda t}$, after $T$ iterations of
Algorithm \ref{alg:residual} we have:
\[
\frac{1}{T} \sum_{t=1}^T [\functional{R}{f_t} - \functional[\textrm{emp}]{R}{f^*}] \le
  \frac{2 c^2 G^2}{\lambda T} (1 + \ln T + \frac{2}{T}).
\]
\end{theorem}

\begin{theorem}
\label{theorem:residual-empirical}
Let $\mathcal{R}_{\textrm{emp}}$ be a convex
functional over $\mathcal{F}$.
Let $\mathcal{H} \subset \mathcal{F}$ be a restriction set
with edge $\gamma$.  Let $\norm[\hat{P}]{\nabla \functional{R}{f}} \le G$
and $\norm[\hat{P}]{f} \le F$ for all $f \in \mathcal{F}$.
Let $f^* = {\argmin}_{f \in \mathcal{F}} \functional[\textrm{emp}]{R}{f}$.
Let $c = \frac{2}{\gamma^2}$.
Given a starting point $f_0$ and
step size $\eta_t = \frac{1}{\sqrt{t}}$, after $T$ iterations of
Algorithm \ref{alg:residual} we have:
\[
\frac{1}{T} \sum_{t=1}^T [\functional[\textrm{emp}]{R}{f_t} - \functional[\textrm{emp}]{R}{f^*}] \le
  \frac{F^2}{2 \sqrt{T}} + \frac{c^2 G^2}{\sqrt{T}} + \frac{c^2 G^2}{2 T^{\frac{3}{2}}}.
\]
\end{theorem}

Again, the results are similar bounds to those from the non-restricted
case.  Like the previous proof, the extra terms in the bound come from
the penalty paid in projection errors at each time step, but here the residual
serves as a mechanism for pushing the error back to later projections.  The
analysis relies on a bound on the norm
of the residual $\Delta$, derived by observing that it is increased by at
most the norm of the gradient and then multiplicatively decreased in projection due to
the edge requirement.  This bound on the size of the residual presents
itself in the $c$ term present in the bound.  Complete proofs are presented in
\ifthenelse{\boolean{longversion}}
{Appendix \ref{app:general}.}
{the extended version of the paper \cite{withproofs}.}

In terms of efficiency, these two algorithms are similarly matched.
For the strongly convex case, the
repeated projection algorithm uses $O(T^2)$ weak learners to obtain
an average regret $O(\frac{\ln T}{T} + \frac{1}{\gamma^2 T})$,
while the
residual algorithm uses $O(T)$ weak learners and has average regret
$O(\frac{\ln T}{\gamma^4 T})$.   The major difference
lies in frequency of the gradient evaluation, where the repeated projection algorithm
evaluates the gradient much less often than the than the residual algorithm.

\section{Experimental Results}

We present preliminary experimental results for these new algorithms on three tasks:
an imitation learning problem, a ranking problem and a set of sample
classification tasks.

The first experimental setup is an optimization problem which
results from the Maximum Margin Planning \cite{ratliff:09a}
approach to imitation learning.
In this setting, a demonstrated policy is provided as example behavior
and the goal is to learn a cost function over features of the environment
which produce policies with similar behavior.  This is done by optimizing over
a convex, non-smooth loss function which minimizes the difference in
costs between the current and demonstrated behavior.  Previous attempts in
the literature have been made to adapt boosting to this setting
\cite{ratliff:09a, bradley:09}, similar to the naive algorithm
presented here, but no convergence results for this settings are known.

\begin{figure}[tb]
\centering
\includegraphics[width=\linewidth]{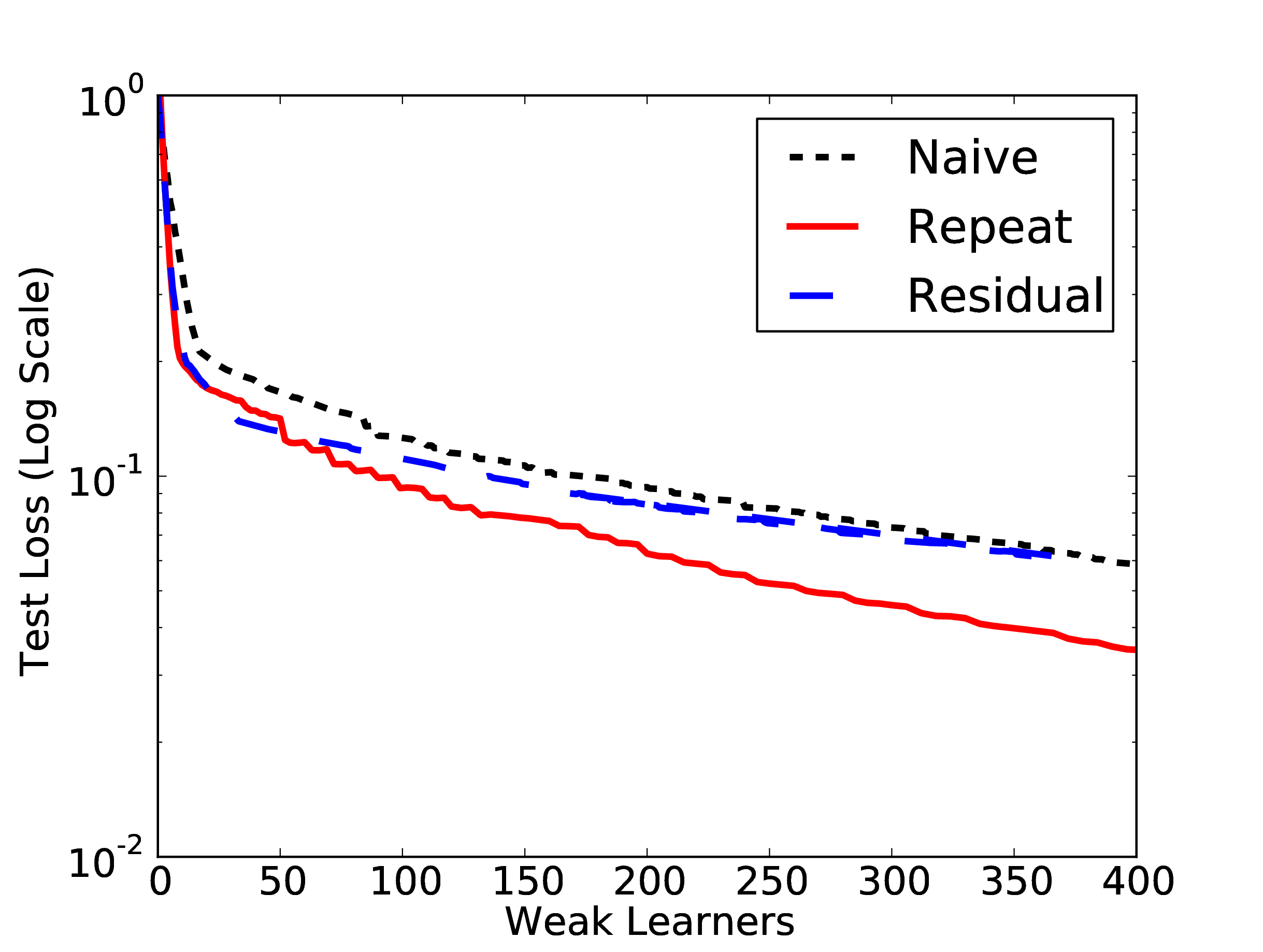}
\caption{Test set loss vs number of weak learners used
for a maximum margin structured imitation
learning problem for all three restricted gradient algorithms.}
\label{fig:mmp-objective}
\end{figure}

\begin{figure}[tb]
\centering
\includegraphics[width=\linewidth]{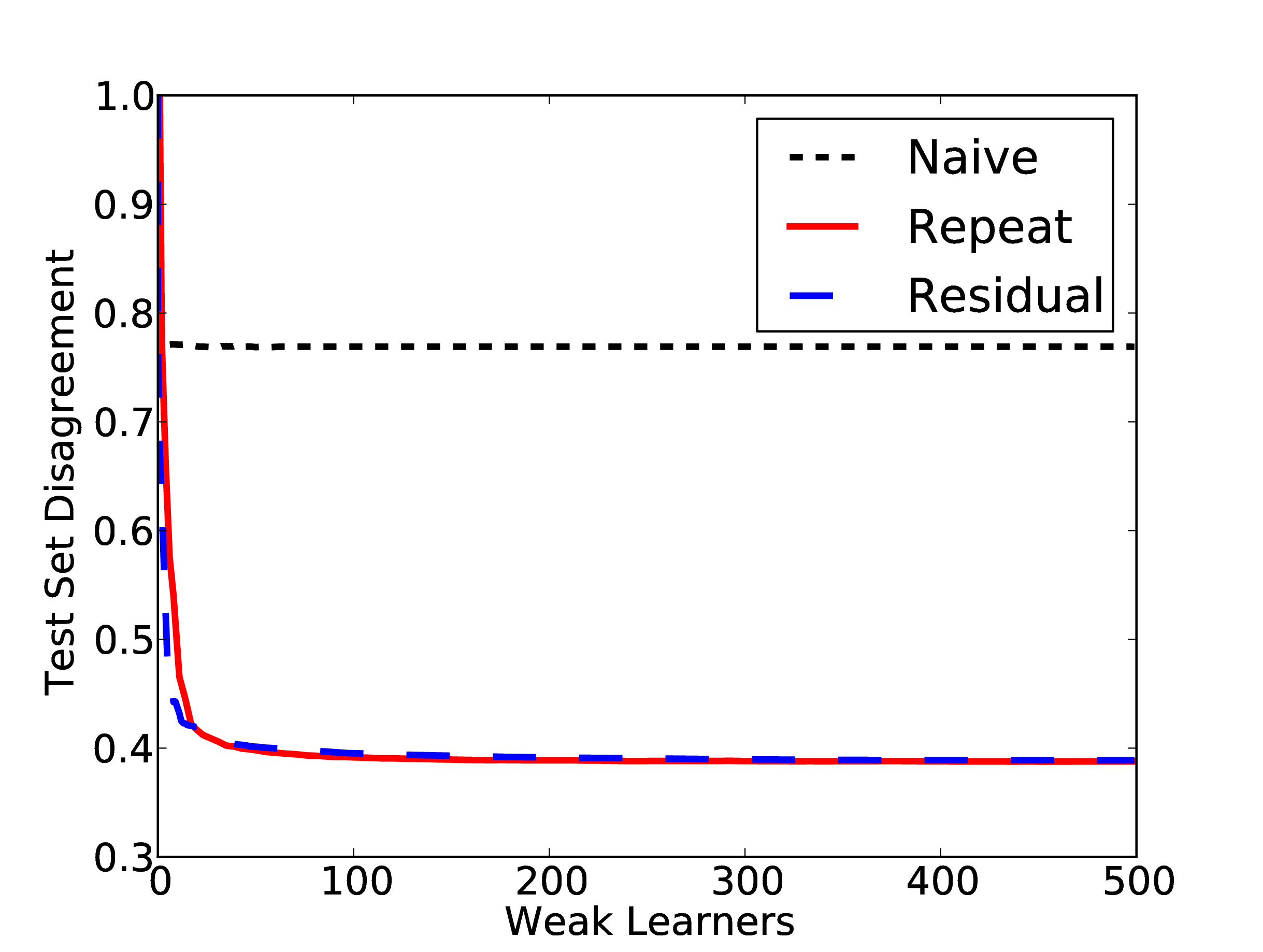}
\caption{Test set disagreement (fraction of violated constraints) vs number of weak learners used
for the MSLR-WEB10K ranking dataset for all three restricted gradient algorithms.}
\label{fig:ranking}
\end{figure}

Figure \ref{fig:mmp-objective}
shows the results of running all three of the algorithms presented here
on a sample planning dataset from this domain.  The weak learners used were
neural networks with 5 hidden units each.

\begin{figure*}[tb]
\centering
\includegraphics[width=0.245\textwidth]{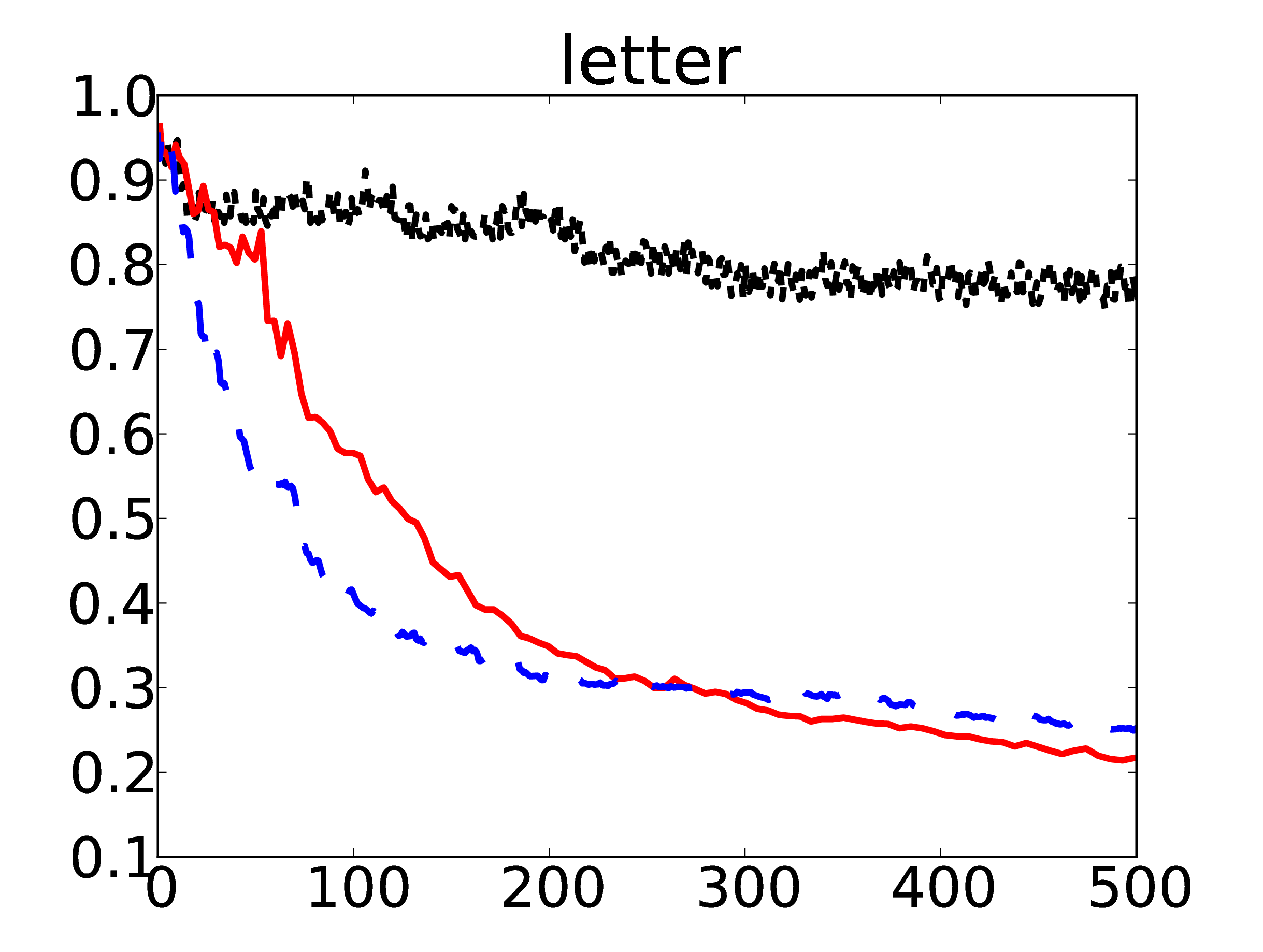}
\includegraphics[width=0.245\textwidth]{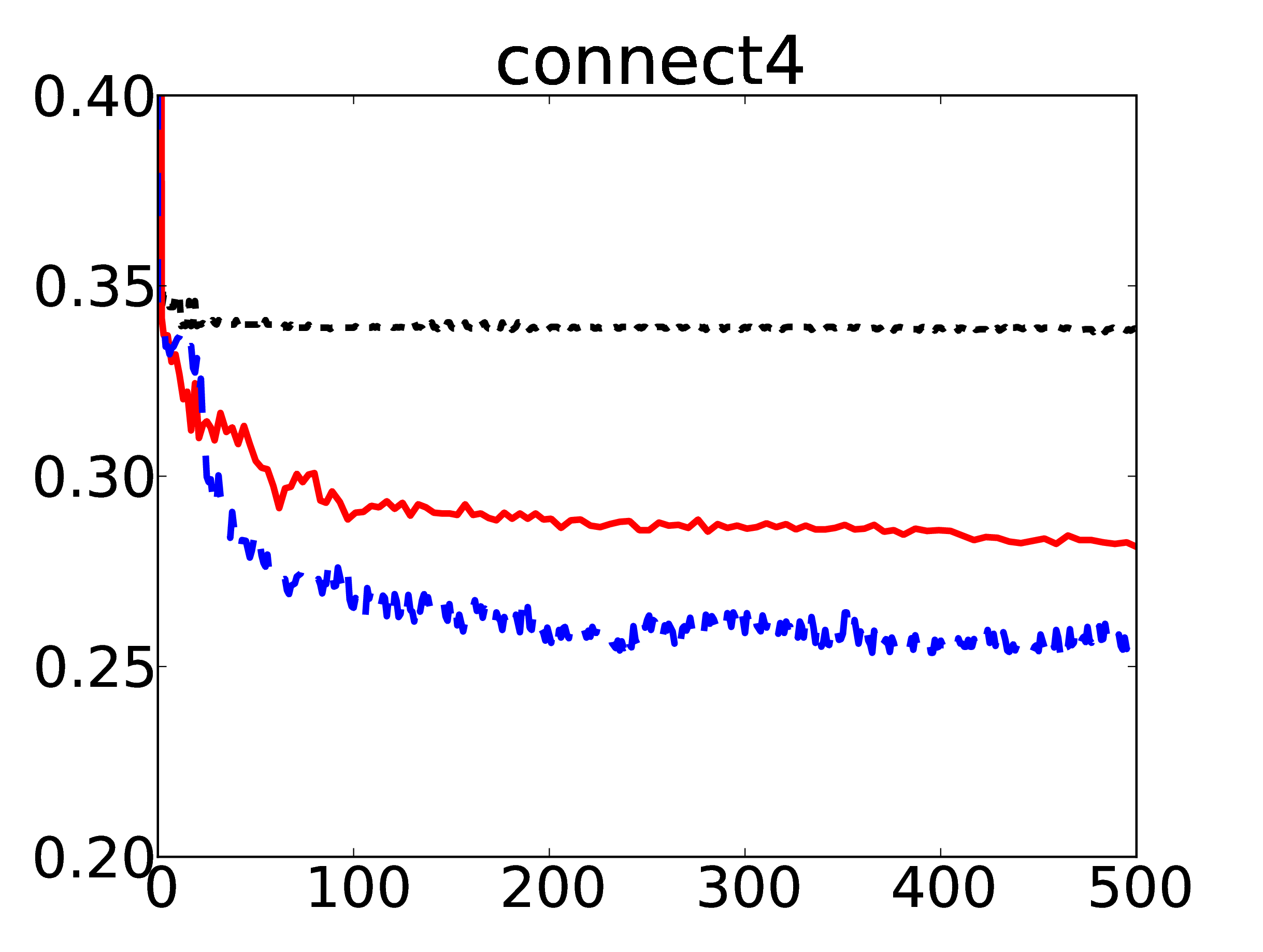}
\includegraphics[width=0.245\textwidth]{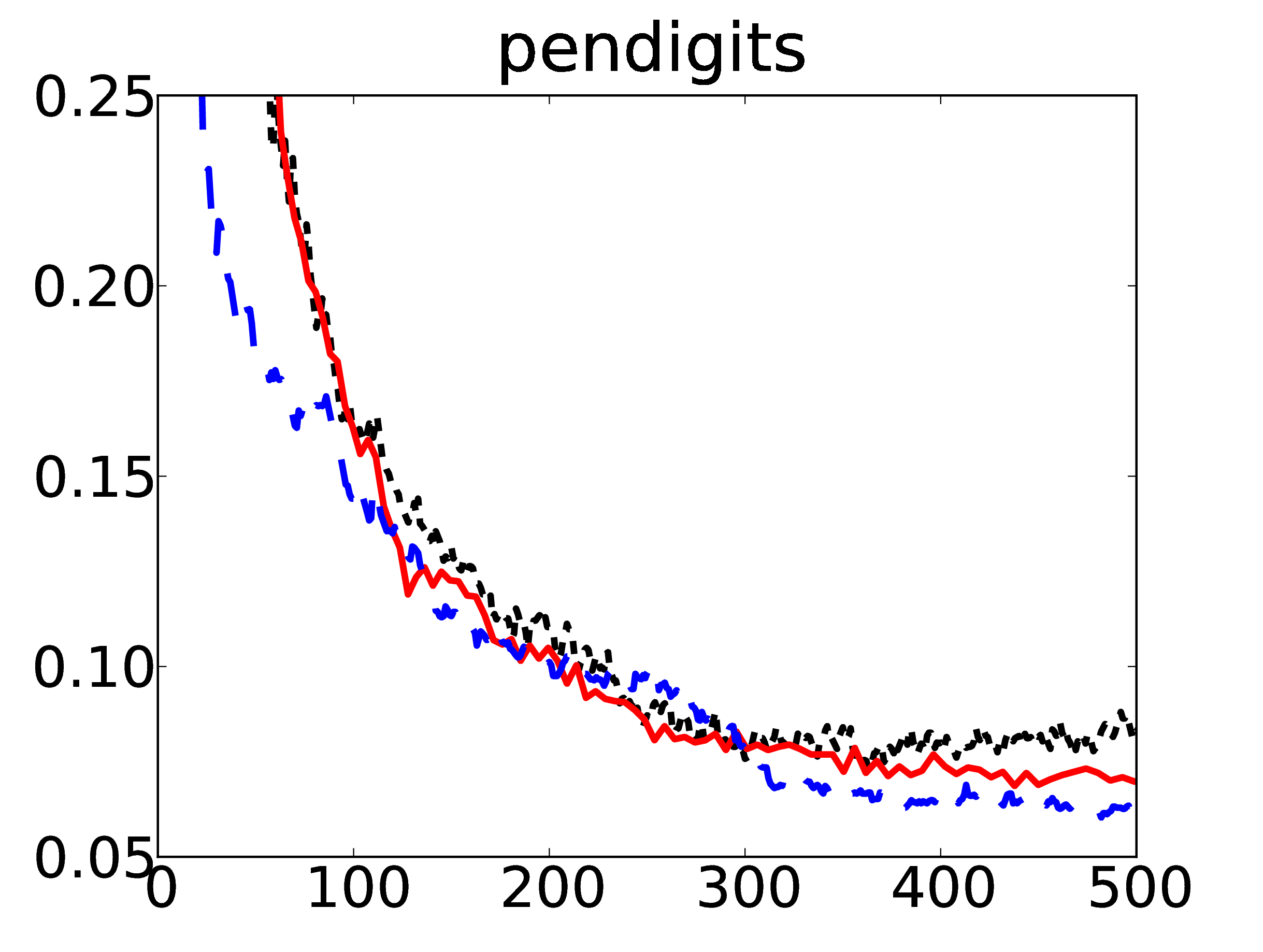}
\includegraphics[width=0.245\textwidth]{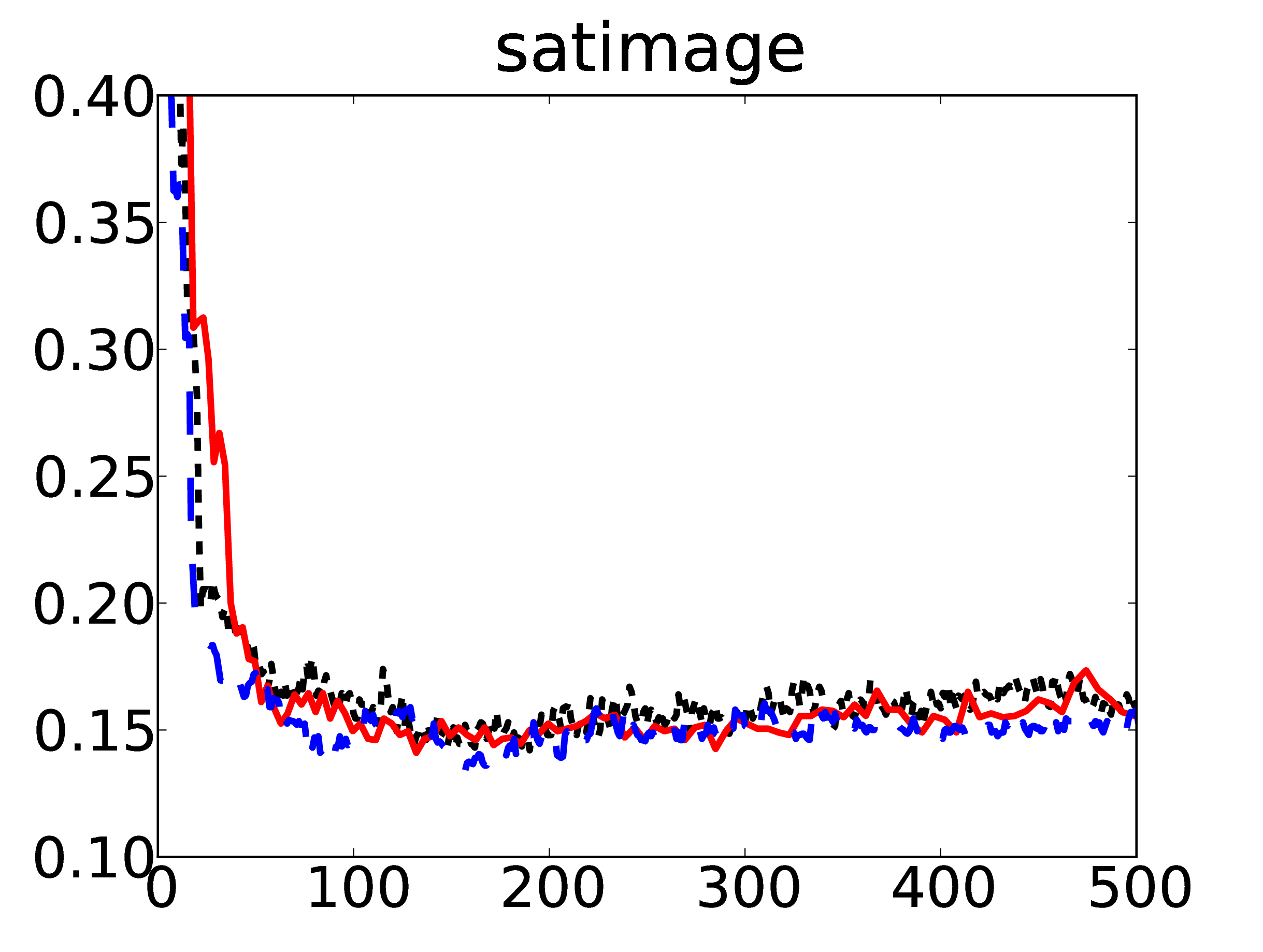}
\caption{Performance on multiclass classification experiments
over the UCI `connect4', `letter', `pendigits' and `satimage' datasets.
The algorithms shown are the
naive projection (black dashed line), repeated projection steps (red solid line),
and the residual projection algorithm (blue long dashed line).}
\label{fig:classification}
\end{figure*}

The second experimental setting is a ranking task from the Microsoft Learning
to Rank Datasets, specifically MSLR-WEB10K \cite{ms:2010},
using the ranking version of the
hinge loss and decision stumps as weak learners.
Figure \ref{fig:ranking} shows the
test set disagreement (the percentage of violated ranking constraints) plotted
against the number of weak learners.

As a final test, we ran our boosting algorithms on several
multiclass classification tasks from the UCI Machine Learning Repository \cite{frank:2010},
using the `connect4', `letter', `pendigits' and `satimage' datasets.
All experiments used the
multiclass extension to the hinge loss \cite{crammer:02}, along
with multiclass decision stumps for the weak learners.

Of particular interest are the experiments
where the naive approach to restricted gradient descent clearly
fails to converge (`connect4' and `letter').  In line with
the presented convergence results, both non-smooth algorithms approach
optimal training performance at relatively similar rates, while the naive
approach cannot overcome the particular conditions of these datasets and
fails to achieve strong performance.  In these cases, the naive approach repeatedly
cycles through the same weak learners, impeding further optimization progress.

{
\section*{Acknowledgements}
We would like to thank Kevin Waugh, Daniel Munoz and the ICML reviewers
for their helpful feedback.  This work was conducted through collaborative
participation in the Robotics Consortium sponsored by the U.S Army
Research Laboratory under the Collaborative Technology Alliance Program,
Cooperative Agreement W911NF-10-2-0016.

\bibliography{references}
\bibliographystyle{icml2011}
}

\onecolumn

\appendixtitleon
\begin{appendices}
\section{Equivalence of boosting requirements}
\label{app:equiv}
First, we demonstrate that our requirement is equivalent to the AdaBoost style
weak learning requirement on weak classifiers.

\begin{reptheorem}{theorem:ada-equivalence}
  For a weak classifier space $\mathcal{H}$
  with outputs in $\{-1,+1\}$, the following statements are
  equivalent: (1) $\mathcal{H}$ has edge $\gamma$ for some
  $\gamma > 0$, and
  (2) for any non-negative weights $w_n$ over training data
  $x_n$, there is a classifier $h \in \mathcal{H}$ which
  achieves an error of at most
  $(\frac{1}{2} - \frac{\delta}{2}) \sum_n w_n$
  for some $\delta > 0$.
\end{reptheorem}

\begin{proof}
To relate the weighted classification setting
and our inner product formulation, let weights $w_n = | \nabla(x_n) |$ and labels
$y_n = \operatorname{sgn}(\nabla(x_n))$.  We examine classifiers $h$ with outputs
in $\{-1,+1\}$.

Consider the AdaBoost weak learner requirement re-written
as a sum over the correct examples:
\[
\sum_{n,h(x_n) = y_n} w_n \ge (\frac{1}{2} + \frac{\delta}{2}) \sum_n w_n.
\]

Breaking the sum over weights into the sum of correct and incorrect weights:
\[
\frac{1}{2} (\sum_{n,h(x_n) = y_n} w_n - \sum_{n,h(x_n) \ne y_n} w_n) \ge \frac{\delta}{2} \sum_n w_n.
\]

The left hand side of this inequality is just $N$ times the inner product $\innerprod{\nabla}{h}$,
and the right hand side can be re-written as the 1-norm of the weight vector $w$, giving:
\begin{align*}
N \innerprod{\nabla}{h} &\ge \delta \norm[1]{w}\\
                        &\ge \delta \norm[2]{w}
\end{align*}

Finally, using $\norm{h} = 1$ and $\norm{\nabla}^2 = \frac{1}{N} \norm[2]{w}^2$:
\[
\innerprod{\nabla}{h} \ge \frac{\delta}{\sqrt{N}} \norm{\nabla} \norm{h}
\]
showing that the AdaBoost requirement implies our requirement for
edge $\gamma > \frac{\delta}{\sqrt{N}} > 0$.

We can show the converse by starting with our weak learner requirement
and expanding:
\begin{align*}
\innerprod{\nabla}{h} &\ge \gamma \norm{\nabla} \norm{h}\\
\frac{1}{N} (\sum_{n,h(x_n) = y_n} w_n - \sum_{n,h(x_n) \ne y_n} w_n) &\ge \gamma \norm{\nabla}
\end{align*}

Then, because $\norm{\nabla}^2 = \frac{1}{N} \norm[2]{w}^2$ and $\norm[2]{w} \ge \frac{1}{\sqrt{N}} \norm[1]{w}$
we get:
\begin{align*}
\sum_{n,h(x_n) = y_n} w_n - \sum_{n,h(x_n) \ne y_n} w_n   &\ge \gamma \frac{1}{N} \norm[1]{w}\\
  &\ge \gamma \sum_n w_n\\
\sum_{n,h(x_n) = y_n} w_n &\ge (\frac{1}{2} + \frac{\gamma}{2}) \sum_n w_n,
\end{align*}
giving the final AdaBoost edge requirement.
\end{proof}

In the first part of this proof, the scaling of $\frac{1}{\sqrt{N}}$ shows that
our implied edge weakens as the number of data points increases in relation to the
AdaBoost style edge requirement, an unfortunate but necessary feature.  This
weakening is necessary because our notion of strong learning is much more
general than other boosting frameworks.  In those settings, strong learning
only guarantees that any dataset can be classified with 0 training error,
while our strong learning guarantee gives optimal performance on any convex
loss function.

\begin{reptheorem}{theorem:multiclass-equivalence}
  For a weak multiclass classifier space $\mathcal{H}$
  with outputs in $\{1,\ldots,K\}$, let the modified hypothesis space
  $\mathcal{H}'$ contain a hypothesis $h': \mathcal{X} \rightarrow \mathbb{R}^K$
  for each $h \in \mathcal{H}$
  such that $h'(x)_k = 1$ if $h(x) = k$ and $h'(x) = -\frac{1}{K-1}$ otherwise.
  Then, the following statements are
  equivalent: (1) $\mathcal{H'}$ has edge $\gamma$ for some $\gamma > 0$, and
  (2) $\mathcal{H}$ satisfies the performance over baseline requirements
  detailed in Theorem 1 of \cite{mukherjee:10}.
\end{reptheorem}

\begin{proof}
In this section we consider the multiclass extension of the previous setting.  Instead of a weight
vector we now have a matrix of weights $w$ where $w_{nk}$ is the weight or reward for
classifying example $x_n$ as class $k$.  We can simply let weights $w_{nk} = \nabla(x_{nk})$ and use the same
weak learning approach as in \cite{mukherjee:10}.  Given
classifiers $h(x)$ which output a label in $\{1,\ldots,K\}$, we convert to an appropriate weak
learner for our setting by building a function $h'(x)$ which outputs a vector $y \in \mathcal{R}^K$ such that
$y_k = 1$ if $h(x) = k$ and $y_k = -\frac{1}{K-1}$ otherwise.

The equivalent AdaBoost style requirement uses costs $c_{nk} = -w_{nk}$ and minimizes
instead of maximizing, but here we state the weight or reward version of the requirement.
More details on this setting can be found in \cite{mukherjee:10}.  We also make the
additional assumption that $\sum{k} w_{nk} = 0, \forall n$ without loss of generality.
This assumption is fine as we can take a given weight matrix $w$ and modify each
row so it has 0 mean, and still have a valid classification matrix as per \cite{mukherjee:10}.
Furthermore, this modification does not affect the edge over random performance
of a multiclass classifier under their framework.

Again consider the multiclass AdaBoost weak learner requirement re-written
as a sum of the weights over the predicted class for each example:
\[
\sum_{n} w_{nh(x_n)} \ge (\frac{1}{K} - \frac{\delta}{K}) \sum_{n,k} w_{nk} + \delta \sum_n w_{ny_n}
\]
we can then convert the sum over correct labels to the max-norm on weights and multiply through
by $\frac{K}{K-1}$:
\begin{align*}
\sum_{n} w_{nh(x_n)} &\ge \frac{1}{K} \sum_{n,k} w_{nk} - \frac{\delta}{K} \sum_{n,k} w_{nk} + \delta \sum_n w_{ny_n}\\
\frac{K}{K-1} \sum_{n} w_{nh(x_n)} &\ge \frac{1}{K-1} \sum_{n,k} w_{nk}
  + \frac{K}{K-1} (\delta \sum_n \norm[\infty]{w_n} - \frac{\delta}{K} \sum_{n,k} w_{nk})\\
\frac{K}{K-1} \sum_{n} w_{nh(x_n)} - \frac{1}{K-1} \sum_{n,k} w_{nk} &\ge
  \frac{K}{K-1} (\delta \sum_n \norm[\infty]{w_n} - \frac{\delta}{K} \sum_{n,k} w_{nk})
\end{align*}
by the fact that the correct label $y_n = \argmax_k w_{nk}$.

The left hand side of this inequality is just the function space inner product:
\[
N \innerprod{\nabla}{h'} \ge \frac{K}{K-1} (\delta \sum_n \norm[\infty]{w_n} - \frac{\delta}{K} \sum_{n,k} w_{nk}).
\]

Using the fact that $\sum_k w_{nk} = 0$ along with $\norm{\nabla} \le \frac{1}{\sqrt{N}} \sum_n \norm[2]{w_n}$
and $\norm{h'} = \sqrt{\frac{K}{K-1}}$ we can now bound the right hand side:
\begin{align*}
N \innerprod{\nabla}{h'} &\ge
  \frac{K}{K-1} \delta \sum_n \norm[\infty]{w_n}\\
&\ge 
  \frac{K}{K-1} \delta \sum_n \norm[2]{w_n}\\
&\ge
  \frac{K}{K-1} \delta \sqrt{N} \norm{\nabla}\\
&\ge
  \sqrt{\frac{K}{K-1}} \delta \sqrt{N} \norm{\nabla} \norm{h'}\\
\innerprod{\nabla}{h} &\ge   \sqrt{\frac{K}{K-1}} \delta \frac{1}{\sqrt{N}} \norm{\nabla} \norm{h'}
\end{align*}

For $K \ge 2$ we get $\gamma \ge \frac{\delta}{\sqrt{N}}$,
showing that the existence of the AdaBoost style edge implies
the existence of ours.  Again, while the requirements
are equivalent for some fixed dataset, we see a weaking of the
implication as the dataset grows large, an unfortunate
consequence of our broader strong learning goals.

Now to show the other direction, start with the inner product formulation:
\begin{align*}
\innerprod{\nabla}{h'} &\ge \delta \norm{\nabla} \norm{h'}\\
\frac{1}{N} (\sum_{n} w_{nh(x_n)} - \frac{1}{K-1} \sum_{n,k \ne h(x_n)} w_{nk}) &\ge \delta \norm{\nabla} \norm{h'}\\
\frac{1}{N} (\frac{K}{K-1} \sum_{n} w_{nh(x_n)} - \frac{1}{K-1} \sum_{n,k} w_{nk}) &\ge \delta \norm{\nabla} \norm{h'}
\end{align*}

Using $\norm{h'} = \sqrt{\frac{K}{K-1}}$ and $\norm{\nabla} \ge \frac{1}{N} \sum_n \norm[2]{w_n}$
we can show:
\[
\frac{K}{K-1} \sum_{n} w_{nh(x_n)} - \frac{1}{K-1} \sum_{n,k} w_{nk}
  \ge \delta \sum_n \norm[2]{w_n} \sqrt{\frac{K}{K-1}}.
\]

Rearranging we get:
\begin{align*}
\frac{K}{K-1} \sum_{n} w_{nh(x_n)}
  &\ge \frac{1}{K-1} \sum_{n,k} w_{nk} + \delta \sum_n \norm[2]{w_n} \sqrt{\frac{K}{K-1}}\\
\sum_{n} w_{nh(x_n)} &\ge \frac{1}{K} \sum_{n,k} w_{nk} + \frac{K-1}{K} \sqrt{\frac{K}{K-1}} \delta \sum_n \norm[2]{w_n}\\
\sum_{n} w_{nh(x_n)} &\ge \frac{1}{K} \sum_{n,k} w_{nk} + \sqrt{\frac{K}{K-1}} \delta (\sum_n \norm[2]{w_n} - \frac{1}{K}\sum_n \norm[2]{w_n})
\end{align*}

Next, bound the 2-norms using $\norm[2]{w_n} \ge \frac{1}{\sqrt{K}} \norm[1]{w_n}$ and $\norm[2]{w_n} \ge \norm[\infty]{w_n}$
and then rewrite as sums of corresponding weights to show the multiclass AdaBoost requirement holds:
\begin{align*}
\sum_{n} w_{nh(x_n)} &\ge (\frac{1}{K} - \frac{\delta}{\sqrt{K-1}K}) \sum_{n,k} w_{nk} + \sqrt{\frac{K}{K-1}} \delta \sum_n \norm[\infty]{w_n}\\
\sum_{n} w_{nh(x_n)} &\ge (\frac{1}{K} - \frac{\delta}{K}) \sum_{n,k} w_{nk} + \delta \sum_n w_{ny_n}\\
\end{align*}

\end{proof}

\section{Smooth Convergence Results}
\label{app:smooth}
For the proofs in this section, all norms and inner products are assumed to be with
respect to the empirical distribution $\hat{P}$.

\begin{reptheorem}{theorem:smooth-empirical}
  Let $\mathcal{R}_{\textrm{emp}}$ be a $\lambda$-strongly convex and
  $\Lambda$-strongly smooth functional over
  $L^2(\mathcal{X},\hat{P})$ space.
  Let $\mathcal{H} \subset L^2$ be a restriction set
  with edge $\gamma$.
  Let $f^* = {\argmin}_{f \in \mathcal{F}} \functional[\textrm{emp}]{R}{f}$.
  Given a starting point $f_0$ and
  step size $\eta_t = \frac{1}{\Lambda}$,
  after $T$ iterations of Algorithm \ref{alg:naive} we have:
  \[
  \functional[\textrm{emp}]{R}{f_T} - \functional[\textrm{emp}]{R}{f^*}
  \le (1 - \frac{\gamma^2 \lambda}{\Lambda})^T
  (\functional[\textrm{emp}]{R}{f_0} - \functional[\textrm{emp}]{R}{f^*}).
  \]
\end{reptheorem}

\begin{proof}
Starting with the definition of strong smoothness, and examining the objective value
at time $t+1$ we have:
\[
\functional{R}{f_{t+1}} \le \functional{R}{f_t}
  + \innerprod{\nabla \functional{R}{f_t}}{f_{t+1} - f_t}
  + \frac{\Lambda}{2} \norm{f_{t+1} - f_t}^2
\]
Then, using $f_{t+1} = \frac{1}{\Lambda}\frac{\innerprod{\nabla \functional{R}{f_t}}{h_t}}{\norm{h_t}^2} h_t$
we get:
\[
\functional{R}{f_{t+1}} \le \functional{R}{f_t}
  - \frac{1}{2 \Lambda} \frac{\innerprod{\nabla \functional{R}{f_t}}{h_t}^2}{\norm{h_t}^2}
\]

Subtracting the optimal value from both sides and applying the edge requirement we get:
\[
\functional{R}{f_{t+1}} - \functional{R}{f^*} \le \functional{R}{f_t} - \functional{R}{f^*}
  - \frac{\gamma}{2 \Lambda} \norm{\nabla \functional{R}{f_t}}^2
\]

From the definition of strong convexity we know $\norm{\nabla \functional{R}{f_t}}^2 \ge 2 \lambda (\functional{R}{f_t} - \functional{R}{f^*})$
where $f^*$ is the minimum point.  Rearranging we can conclude that:
\[
\functional{R}{f_{t+1}} - \functional{R}{f^*} \le (\functional{R}{f_t} - \functional{R}{f^*})(1 - \frac{\gamma \lambda}{\Lambda})
\]

Recursively applying the above bound starting at $t=0$ gives the final bound on $\functional{R}{f_T} - \functional{R}{f_0}$.
\end{proof}

\section{General Convergence Results}
\label{app:general}
For the proofs in this section, all norms and inner products are assumed to be with
respect to the empirical distribution $\hat{P}$.

\begin{reptheorem}{theorem:repeat-empirical-strong-convex}
  Let $\mathcal{R}_{\textrm{emp}}$ be a $\lambda$-strongly convex
  functional over $\mathcal{F}$.
  Let $\mathcal{H} \subset \mathcal{F}$ be a restriction set
  with edge $\gamma$.  Let $\norm[\hat{P}]{\nabla \functional{R}{f}} \le G$.
  Let $f^* = {\argmin}_{f \in \mathcal{F}} \functional[\textrm{emp}]{R}{f}$.
  Given a starting point $f_0$ and
  step size $\eta_t = \frac{2}{\lambda t}$, after $T$ iterations of
  Algorithm~\ref{alg:repeat} we have:
  \[
  \frac{1}{T} \sum_{t=1}^T [\functional[\textrm{emp}]{R}{f_t} - \functional[\textrm{emp}]{R}{f^*}] \le
  \frac{G^2}{\lambda T} (1 + \ln T + \frac{1 - \gamma^2}{\gamma^2}).
  \]
\end{reptheorem}

\begin{proof}
First, we start by bounding the potential $\norm{f_{t} - f^*}^2$,
similar to the potential function arguments in \cite{zinkevich:03,hazan:06},
but with a different descent step:
\begin{align*}
\norm{f_{t+1} - f^*}^2 &\le \norm{f_t - \eta_t (h_t) - f^*}^2\\
&= \norm{f_t - f^*}^2 + \eta_t^2 \norm{h_t}^2 - 2 \eta_t \innerprod{f_t - f^*}{h_t - \nabla_t} - 2 \eta_t \innerprod{f_t - f^*}{\nabla_t}\\
\innerprod{f^* - f_t}{\nabla_t} &\le \frac{1}{2\eta_t} \norm{f_{t+1} - f^*}^2 -
  \frac{1}{2\eta_t} \norm{f_t - f^*}^2 - 
  \frac{\eta_t}{2} \norm{h_t}^2 - \innerprod{f^* - f_t}{h_t - \nabla_t}
\end{align*}

Using the definition of strong convexity and summing:
\begin{align*}
\sum_{t=1}^T \risk{f^*} &\ge \sum_{t=1}^T \risk{f_t} +
  \sum_{t=1}^T \innerprod{f^* - f_t}{\nabla_t} + \sum_{t=1}^T \frac{\lambda}{2} \norm{f^* - f_t}^2\\
\begin{split}
&\ge \sum_{t=1}^T \risk{f_t} - \frac{1}{\eta_1} \norm{f_1 - f^*}^2 +
  \sum_{t=1}^{T-1} \frac{1}{2} \norm{f_{t+1} - f^*}^2 (\frac{1}{\eta_{t}} - \frac{1}{\eta_{t+1}} + \lambda) -\\
&\quad
  \sum_{t=1}^T \frac{\eta_t}{2} \norm{h_t}^2 -
  \sum_{t=1}^T \innerprod{f^* - f_t}{h_t - \nabla_t}
\end{split}\\
\end{align*}

Setting $\eta_t = \frac{2}{\gamma t}$ and use bound $\norm{h_t} \le 2 \norm{\nabla_t} \le 2 G$ :
\begin{align*}
\sum_{t=1}^T \risk{f^*}
&\ge \sum_{t=1}^T \risk{f_t} -
  \frac{4 G^2}{2} \sum_{t=1}^T \frac{2}{\lambda t} -
  \frac{\lambda}{4} \sum_{t=1}^T (\norm{f_t - f^*}^2 -
  \sum_{t=1}^T \innerprod{f^* - f_t}{h_t - \nabla_t})\\
&\ge \sum_{t=1}^T \risk{f_t} -
  \frac{4 G^2}{\lambda} (1 + \ln T) -
  \frac{1}{\lambda} \sum_{t=1}^T \norm{h_t - \nabla_t}^2
\end{align*}

Using the result from \ref{theorem:smooth-empirical} we can bound the error at each step $t$:
\begin{align*}
\sum_{t=1}^T \risk{f^*}
&\ge \sum_{t=1}^T \risk{f_t} -
  \frac{4 G^2}{\lambda} (1 + \ln T) -
  \frac{G^2}{\lambda} \sum_{t=1}^T (1 - \gamma^2)^t\\
&\ge \sum_{t=1}^T \risk{f_t} -
  \frac{4 G^2}{\lambda} (1 + \ln T) -
  \frac{G^2}{\lambda} \frac{1 - \gamma^2}{\gamma^2}
\end{align*}
giving the final bound.
\end{proof}

\begin{reptheorem}{theorem:repeat-empirical}
Let $\mathcal{R}_{\textrm{emp}}$ be a convex
functional over $\mathcal{F}$.
Let $\mathcal{H} \subset \mathcal{F}$ be a restriction set
with edge $\gamma$.  Let $\norm[\hat{P}]{\nabla \functional{R}{f}} \le G$
and $\norm[\hat{P}]{f} \le F$ for all $f \in \mathcal{F}$.
Let $f^* = {\argmin}_{f \in \mathcal{F}} \functional[\textrm{emp}]{R}{f}$.
Given a starting point $f_0$ and
step size $\eta_t = \frac{1}{\sqrt{t}}$, after $T$ iterations of
Algorithm \ref{alg:repeat} we have:
\[
\frac{1}{T} \sum_{t=1}^T [\functional[\textrm{emp}]{R}{f_t} - \functional[\textrm{emp}]{R}{f^*}] \le
  \frac{F^2}{2 \sqrt{T}} + \frac{G^2}{\sqrt{T}} + 2 F G \frac{1-\gamma^2}{\gamma^2}.
\]
\end{reptheorem}

\begin{proof}
Like the last proof, we start with the altered potential and sum over the definition of
convexity:
\begin{align*}
\sum_{t=1}^T \risk{f^*}
\ge &\sum_{t=1}^T \risk{f_t} - \frac{1}{\eta_1} \norm{f_1 - f^*}^2 +
  \sum_{t=1}^{T-1} \frac{1}{2} \norm{f_{t+1} - f^*}^2 (\frac{1}{\eta_{t}} - \frac{1}{\eta_{t+1}}) -\\
  &\sum_{t=1}^T \frac{\eta_t}{2} \norm{h_t}^2 -
  \sum_{t=1}^T \innerprod{f^* - f_t}{h_t - \nabla_t}
\end{align*}

Setting $\eta_t = \frac{1}{\sqrt{t}}$ and using bound $\norm{h_t} \le \norm{\nabla_t} \le G$
and the result from \ref{theorem:smooth-empirical} we can bound the error at each step $t$:
\begin{align*}
\sum_{t=1}^T \risk{f^*}
&\ge \sum_{t=1}^T \risk{f_t} - \frac{1}{\eta_T} \norm{f_T - f^*}^2 -
  \frac{G^2}{2} \sum_{t=1}^T \frac{1}{\sqrt{t}} - \sum_{t=1}^T \innerprod{f^* - f_t}{h_t - \nabla_t}\\
&\ge \sum_{t=1}^T \risk{f_t} - \frac{F^2 \sqrt{T}}{2} - G^2 \sqrt{T} -
  F G \sum_{t=1}^T \sqrt{(1 - \gamma^2)^t}\\
&\ge \sum_{t=1}^T \risk{f_t} - \frac{F^2 \sqrt{T}}{2} - G^2 \sqrt{T} -
  2 F G \frac{1 - \gamma^2}{\gamma^2}
\end{align*}
giving the final bound.
\end{proof}

\begin{reptheorem}{theorem:residual-empirical-strong-convex}
Let $\mathcal{R}_{\textrm{emp}}$ be a $\lambda$-strongly convex
functional over $\mathcal{F}$.
Let $\mathcal{H} \subset \mathcal{F}$ be a restriction set
with edge $\gamma$.  Let $\norm[\hat{P}]{\nabla \functional{R}{f}} \le G$.
Let $f^* = {\argmin}_{f \in \mathcal{F}} \functional[\textrm{emp}]{R}{f}$.
Let $c = \frac{2}{\gamma^2}$.
Given a starting point $f_0$ and
step size $\eta_t = \frac{1}{\lambda t}$, after $T$ iterations of
Algorithm \ref{alg:residual} we have:
\[
\frac{1}{T} \sum_{t=1}^T [\functional{R}{f_t} - \functional[\textrm{emp}]{R}{f^*}] \le
  \frac{2 c^2 G^2}{\lambda T} (1 + \ln T + \frac{2}{T}).
\]
\end{reptheorem}

\begin{proof}
Like the proof of Theorem \ref{theorem:repeat-empirical-strong-convex},
we again use a potential function and sum over the definition of convexity:
\begin{align*}
\sum_{t=1}^T \risk{f^*} 
\ge &\sum_{t=1}^T \risk{f_t} - \frac{1}{\eta_1} \norm{f_1 - f^*}^2 +
  \sum_{t=1}^{T-1} \frac{1}{2} \norm{f_{t+1} - f^*}^2 (\frac{1}{\eta_{t}} - \frac{1}{\eta_{t+1}} + \lambda) -\\
  &\sum_{t=1}^T \frac{\eta_t}{2} \norm{h_t}^2 -
  \sum_{t=1}^T \innerprod{f^* - f_t}{h_t - (\Delta_t + \nabla_t)} - \sum_{t=0}^{T-1} \innerprod{f^* - f_{t+1}}{\Delta_{t+1}}\\
\ge &\sum_{t=1}^T \risk{f_t} - \frac{1}{\eta_1} \norm{f_1 - f^*}^2 +
  \sum_{t=1}^{T-1} \frac{1}{2} \norm{f_{t+1} - f^*}^2 (\frac{1}{\eta_{t}} - \frac{1}{\eta_{t+1}} + \lambda) -\\
  &\sum_{t=1}^T \frac{\eta_t}{2} \norm{h_t}^2 -
  \sum_{t=1}^T \innerprod{f^* - f_t}{h_t - (\Delta_t + \nabla_t)} - \sum_{t=0}^{T-1} \innerprod{f^* - f_t}{\Delta_{t+1}} -
  \sum_{t=0}^{T-1} \innerprod{\eta_{t} h_t}{\Delta_{t+1}}
\end{align*}
where $h_t$ is the augmented step taken in Algorithm \ref{alg:residual}.

Setting $\eta_t = \frac{1}{\gamma t}$ and use bound $\norm{h_t} \le \norm{\nabla_t} \le G$,
along with $\Delta_{t+1} = (\Delta_t + \nabla_t) - h_t$:
\begin{align*}
\sum_{t=1}^T \risk{f^*}
&\ge \sum_{t=1}^T \risk{f_t} \sum_{t=1}^T \frac{\eta_t}{2} \norm{h_t}^2 -
  (\innerprod{f^* - f_{T+1}}{\Delta_{t+1}} - \frac{\lambda T}{2} \norm{f^*-f_{T+1}}^2) -
  \sum_{t=1}^{T} \innerprod{\eta_{t} h_t}{\Delta_{t+1}}
\end{align*}

We can bound the norm of $\Delta_t$ by considering that (a) it start at $0$ and (b) at
each time step it increases by at most $\nabla_t$ and is multiplied by $1 - \gamma^2$.
This implies that $\norm{\Delta_t} \le c G$ where $c = \frac{\sqrt{1-\gamma^2}}{1 - \sqrt{1 - \gamma^2}} < \frac{2}{\gamma^2}$.

From here we can get a final bound:
\begin{align*}
\sum_{t=1}^T \risk{f^*}
\ge &\sum_{t=1}^T \risk{f_t} - \frac{c^2 G^2}{\lambda} (1 + \ln T) -
  \frac{2 c^2 G^2}{\lambda T} -
  \frac{c^2 G^2}{\lambda} (1 + \ln T)
\end{align*}
\end{proof}

\begin{reptheorem}{theorem:residual-empirical}
Let $\mathcal{R}_{\textrm{emp}}$ be a convex
functional over $\mathcal{F}$.
Let $\mathcal{H} \subset \mathcal{F}$ be a restriction set
with edge $\gamma$.  Let $\norm[\hat{P}]{\nabla \functional{R}{f}} \le G$
and $\norm[\hat{P}]{f} \le F$ for all $f \in \mathcal{F}$.
Let $f^* = {\argmin}_{f \in \mathcal{F}} \functional[\textrm{emp}]{R}{f}$.
Let $c = \frac{2}{\gamma^2}$.
Given a starting point $f_0$ and
step size $\eta_t = \frac{1}{\sqrt{t}}$, after $T$ iterations of
Algorithm \ref{alg:residual} we have:
\[
\frac{1}{T} \sum_{t=1}^T [\functional[\textrm{emp}]{R}{f_t} - \functional[\textrm{emp}]{R}{f^*}] \le
  \frac{F^2}{2 \sqrt{T}} + \frac{c^2 G^2}{\sqrt{T}} + \frac{c^2 G^2}{2 T^{\frac{3}{2}}}.
\]
\end{reptheorem}

\begin{proof}
Similar to the last few proofs, we get a result similar to the standard gradient version, with
the error term from the last proof:
\begin{align*}
\sum_{t=1}^T \risk{f^*}
\ge &\sum_{t=1}^T \risk{f_t} - \frac{1}{\eta_1} \norm{f_1 - f^*}^2 +
  \sum_{t=1}^{T-1} \frac{1}{2} \norm{f_{t+1} - f^*}^2 (\frac{1}{\eta_{t}} - \frac{1}{\eta_{t+1}}) -\\
  &\sum_{t=1}^T \frac{\eta_t}{2} \norm{h_t}^2 -
  (\innerprod{f^* - f_{T+1}}{\Delta_{t+1}} - \frac{\sqrt{T}}{2} \norm{f^*-f_{T+1}}^2) -
  \sum_{t=1}^{T} \innerprod{\eta_{t} h_t}{\Delta_{t+1}}
\end{align*}

Using the bound on $\norm{\Delta_t} \le c$ from above and setting $\eta_t = \frac{1}{\sqrt{t}}$:
\begin{align*}
\sum_{t=1}^T \risk{f^*}
&\ge \sum_{t=1}^T \risk{f_t} - \frac{F^2 \sqrt{T}}{2} - c^2 G^2 \sqrt{T} -
  \frac{c^2 G^2}{2 \sqrt{T}}
\end{align*}
giving the final bound.
\end{proof}

\end{appendices}

\end{document}